# Assessing Geo-Foundational Models for Flood Inundation Mapping: Benchmarking models for Sentinel-1, Sentinel-2, and Planetscope

Saurabh Kaushik, Lalit Maurya, Elizabeth Tellman and ZhiJie Zhang

*Abstract*— **Geo-Foundational Models (GFMs) enable fast and reliable extraction of spatiotemporal information from satellite imagery, improving flood inundation mapping by leveraging location and time embeddings. Despite their potential, it remains unclear whether GFMs outperform traditional models like U-Net. A systematic comparison across sensors and data availability scenarios is still lacking, which is essential to guide end-users in model selection. To address this, we evaluate three GFMs—Prithvi 2.0, Clay V1.5, DOFA—and UViT (a Prithvi variant), against TransNorm, U-Net, DeepLabv3+, and Attention U-Net using PlanetScope, Sentinel-1, and Sentinel-2. We observe competitive performance among all GFMs, with 2–5% variation between the best and worst models across sensors. Clay outperforms others on PlanetScope (0.79 mIoU) and Sentinel-2 (0.72), while Prithvi leads on Sentinel-1 (0.57). In leave-one-region-out cross-validation across five regions, Clay shows slightly better performance across all sensors, with mIoU scores of 0.72, 0.66, and 0.51 for PanetScope, Sentinel-2, and Sentinel-1 respectively, compared to Prithvi (0.70, 0.64, 0.49) and DOFA (0.67, 0.64, 0.49).Across all 19 sites, cross-validation reveals a 4% improvement by Clay over U-Net. Visual inspection highlights Clay's superior ability to retain fine details. Few-shot experiments show Clay achieves 0.64 mIoU on PlanetScope with just five training images, outperforming Prithvi (0.24) and DOFA (0.35). In terms of computational time, Clay is a better choice due to its smaller model size (26M parameters), making it approximately 3× faster than Prithvi (650M) and 2× faster than DOFA (410M). Our results suggest GFMs offer small to moderate improvements in flood mapping accuracy at lower computational cost and labeling effort.**

***Index Terms*—Geo-Foundation Models, Prithvi, Clay, DOFA, Vision Transformers, CNN, Flood, Remote Sensing.**

## I. INTRODUCTION

Floods impact millions of people globally, with exposed population increase due to climate and demographic changes [1], [2]. The economic impact of floods can be witnessed globally, and in the USA alone, loses due to floods are estimated between 179.8 and 496.0 billion each year, which are anticipated to increase in the near future [3]. The direct exposure of population to floods is particularly high in low-income countries, South and East Asia alone accounts for 1.24 billion people-about one third of global exposure. In addition, 170 million people living in extreme poverty (less than USD$1.90/day) reside in high flood-risk areas [4]. To mitigate the impact of floods, fast and accurate flood maps play an imperative role, to improve emergency response [5], target flood recovery efforts [6], and develop new types of flood insurance [7].

Satellite imagery and recent developments in deep learning algorithms, accompanied by improved computational power, offer pragmatic solutions to produce fast and reliable flood maps. Several deep learning algorithms such as UNet [8], [9], Attention U-Net [10], DeepLabv3+ [11], have been employed extensively to map floods using satellite data, including combination of Convolutional Neural Network (CNN) and Long-Short Term Memory Network (LSTM) which take advantage of spatial and temporal information (long-range dependencies) to generate long-term flood records [12]. The development of vision transformers which offer state of the art capabilities in retaining long-range-dependencies has also delivered competitive performance compared to traditional CNN algorithms in computer vision tasks, including image segmentation [13], [14] and flood mapping specifically [15]. These models have the advantage of considering long-range spatial (global) dependencies along with retaining local information. Recent work in earth observation has focused on leveraging the capabilities of CNN and ViT using hybrid models, where ViT encoders are used to capture long- range dependencies and CNN decoders are used to learn spatial/local characteristics [16], [17], [18]. However, this approach usually requires model training from scratch, which demands high computational resources and a high volume of training data.

To address some of these limitations, vision transformer- based Geo-Foundational Models (GFMs) offer unique capabilities to produce fast and reliable flood inundation maps without building custom models from scratch or requiring high volume of training data [19], [20]. The advantages of GFMs over custom models can be largely attributed to their massive pretrained encoders, which utilize self-supervised learning techniques to leverage vast amounts of unlabeled data. Recently, several GFMs have been introduced, pretrained on diverse sensors using self-supervision approaches. Among these, Masked Autoencoder

This work was funded by the NASA THP Program (award number 80NSSC21K1341). *(Corresponding author: Saurabh Kaushik).* Saurabh Kaushik and Elizabeth Tellman is with Center for Sustainability and the Global Environment (SAGE), University of Wisconsin–Madison, Madison, WI, 53726 USA (e-mail: skaushik8@wisc.edu, beth.tellman@wisc.edu). Lalit Maurya is with School of Computing, University of Portsmouth, Portsmouth, PO1 3HE, UK (e-mail: lalit.maurya@port.ac.uk). ZhiJie Zhang is with Quinney College of Agriculture & Natural Resources, Department of Environment and Society, Utah State University, Logan, USA (e-mail: zhijie.zhang@usu.edu)

(MAE) [21]-based models have gained significant adoption [19], [22], [23], [24], while others are built on contrastive learning frameworks [25], [26], [27]. The primary goal of these foundational models is to achieve generalization across space and time for a wide range of Earth observation tasks, including semantic segmentation, classification, object detection, and regression. The overarching vision is to develop a single model that can be fine-tuned for multiple downstream tasks, rather than training application-specific models from scratch. To evaluate their performance, the community has relied on two major benchmark datasets: PANGEA [28] and Geo-Bench [29]. Many of these models [19], [24], [25] have demonstrated promising results for flood-inundation mapping as a downstream task using the Sen1Floods11 dataset [30].

The recent evaluations benchmarking 12 geo-foundational models for flood mapping on the Sentinel-2 sensor using Sen1Floods11 [28] found that conventional U-Net outperformed all of the GFM. However, the evaluations show competitive performance from Contrastive Radar-Optical Masked Autoencoders (CROMA) (mIoU = 90.89) (Fuller et al., 2023) and TerraMind (mIoU = 90.78), compared to the benchmark U-Net (mIoU = 91.42). The most interesting observation from detailed analysis revealed that U-Net outperformed all GFMs consistently even in limited data scenarios (10%, 50% and 100%). These results have sparked an ongoing debate within the scientific community regarding the practical utility of large pretrained encoders and advanced Vision Transformer (ViT)-based architectures, which demand significantly higher computational resources and memory compared to traditional CNN-based models. However, we would like to emphasize that benchmarking evaluation of Sen1Floods11 included frozen encoder weights and only decoder (UperNet) is trained to generate segmentation mask. Thereby, full potential of GFMs with full fine-tuning and hyperparameter tuning is yet to be unlocked. In addition, application of existing GFMs over commercial data such as PlanetScope, inter-sensor comparison, and labeling effort required to produce good results has not yet been evaluated and are core questions for end-users seeking to make the best flood map possible in an operational setting (e.g. for disaster response, relief, or insurance). Unlike previous work, we examine an essential practical question for downstream users- How much labeled data is needed to get the gains from these GFMs on a downstream task like inundation mapping?

This paper compares three foundational models for flood inundation mapping and estimates their relative contribution to common approaches using CNNs or ViTs on labeled datasets trained from scratch. Here we attempt to answer four research questions 1) Do geo-foundational models Prithvi, DOFA, and Clay increase accuracy for flood inundation mapping? 2) Does GFM performance vary across sensors? 3) How does training data size influence GFM accuracy for flood mapping tasks? 4) Which foundational model is ideal in terms of computational resources and limited training data availability? To answer these questions here, we conducted 140 experiments and tested the applicability of Prithvi 2.0 [19], DOFA [24], Clay V1.5 [22], Prithvi variant ”UViT” [31], TransNorm [32] , U-Net [33], and Attention UNet [34] over three sensors, Planetscope, Sentinel-2 and Sentinel-1.

For our experiments we used a new publicly available dataset called FloodPlanet, which covers 19 flood events globally across diverse eco-regions between 2017 and 2020 for all three sensors with high quality, hand labeled data [35]. While other single modality benchmarking datasets exist for Sentinel-1 [36], [37], [38], we sought a dataset that had coincident images across both sensors, and had higher quality labels using Planetscope (a commercial sensor at higher spatial resolution). We provide comparisons of deep learning models and quantify the relative contribution of foundation models across all three sensors. Our findings aid end users in selecting models and datasets for producing fast and reliable flood maps for users to determine trade-offs in time, labeling effort, and computational cost for desired accuracy gains. We selected three foundation models that were trained in both radar and optical sensors, and the only two in our knowledge trained in very high-resolution imagery (Clay: 0.6-1m NAIP, 3m PlanetScope and DOFA: 3m Gaofen). To our knowledge, this research is the first evaluation of Clay's flood inundation mapping abilities. As we expect more foundational models and embeddings are released [39], [40], [41]and existing ones to improve (e.g. future versions of Clay), we hope the FloodPlanet dataset and the evaluation method adopted in this paper can be replicated to assess the relative gains new models can offer for flood mapping tasks. Beyond flood mapping, our results on foundation model fine tuning gives guidance regarding the labeling effort required to fine tune foundation models for other downstream tasks.

## II. Materials and Methods

### *A. Data*

Here we utilized FloodPlanet dataset (https://doi.org/10.25739/m69q-8k22) [35], which is a high quality manually annotated multi-sensor dataset labeled using high resolution (3m) PlanetScope imagery. The dataset covers a total of 19 globally sampled flood events (Fig. 1) and manually labeled inundation during flood events that include flood water and the open water prior to flood. The dataset consists of 366 images of 1024×1024 manually labeled PlanetScope tiles with an extent equivalent to 3500 km$^2$, encompassing eight different types of ecoregions and five continents. It also includes 362 overlapping Sentinel-1 images and 298 Sentinel-2 images that coincide with PlanetScope acquisitions within a 72-hour window (Table I).The dataset was constructed using larger tiles (1024×1024) to incorporate abundant spatial information of inundation. In all experiments, manual labels from high-resolution commercial data (PlanetScope) were used, as initial tests demonstrated improved performance with high-resolution labels compared to lower-resolution labels derived from Sentinel-1 and Sentinel-2. The original published FloodPlanet datasets of Sentinel-1 and Sentinel-2 were of inconsistent size (e.g., 316×316, 324×324 etc.), here we resize these images to 320×320 to have consistent data size and align with Prithvi and Clay patch sizes.

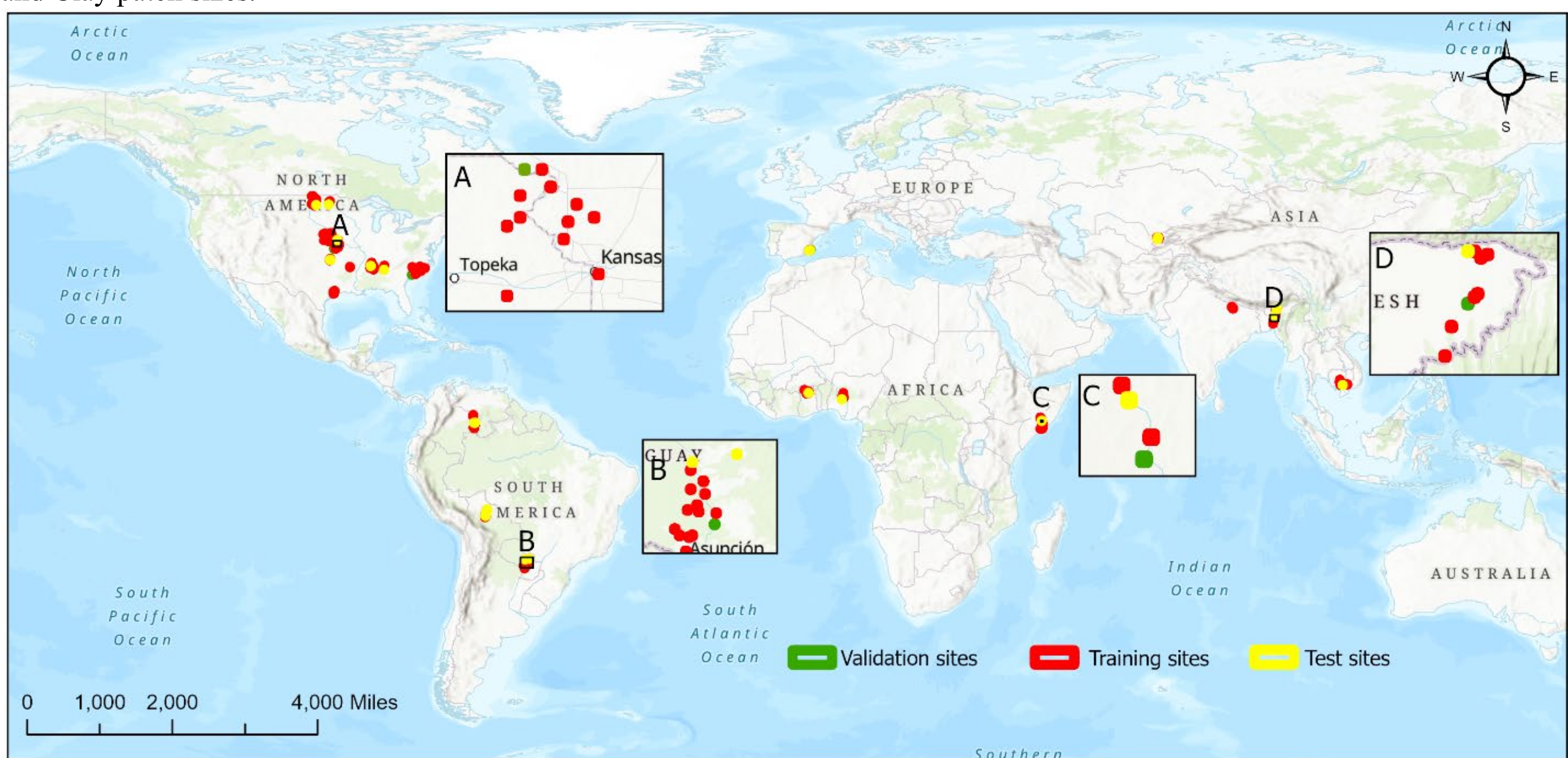

**Fig. 1.** Spatial distribution of training, testing, and validation sites used for evaluating deep learning models. Panels show zoomed-in views of site distribution in parts of (A) North America, (B) South America, (C) Africa, and (D) Asia. Image source: Esri, TomTom.

TABLE I

DATA SPECIFICATION

| Sensors | Dataset Size | Area Coverage |
|---|---|---|
| PlanetScope | 366 (1024×1024) | 3500 |
| Sentinel 1 | 362 (320×320) | 3709 |
| Sentinel 2 | 298 (320×320) | 3051 |

### *B. Models*

The development of large vision foundational models for earth observation have gained wide scientific attention, as we witnessed ~58 GFMs between June 2021 and June 2024 [42]. However, testing each GFM is not practically feasible. Since our objective is to compare GFM performance across sensors for flood mapping, our model selection is guided by three key factors: 1) size of pre-training data, 2) inclusion of multiple sensors including high resolution images, and 3) segmentation results over GEO-Bench [29] and PANGEA [28] benchmark datasets to compare those findings to a higher quality inundation dataset. Based on these factors we selected three GFMs: Prithvi 2.0 [19], DOFA [24], and Clay [22], with approximately 4.2, 8, and 70 million pre-training images, respectively (Table II). Prithvi 2.0 was chosen over other foundational models like DINO [27], DeCUR [26], ScaleMAE [23], and Satlas [43] because either these models ingest only optical data or have been surpassed by Prithvi 2.0 in image segmentation tasks using GEO-Bench dataset [19]. Meanwhile, the Presto [20] model was not selected due to its limitations in mapping small features (e.g., small agriculture filed, streets), particularly when only a small portion of the image contains the target label (e.g., highways or rivers, F1:0.40), which could be a significant drawback in many flood-prone regions. Despite the lack of scientific evaluation of Clay for flood mapping, its distinctive integration of multi-sensor, multi-platform datasets, combined with extensive pretraining on over 70 million image chips sets it apart from other models. DOFA [24] pretrained on 8 million images from diverse sensors and platforms, offers unique capabilities for integrating remote sensing data of any number of channels. Additionally, DFOA is an ideal comparative model for Clay considering its pretraining over high-resolution datasets such as Gaofen-2 (3m) and NAIP (0.6-1m). Given these factors, we selected Prithvi, DOFA, and Clay for a focused and representative evaluation in flood mapping, acknowledging that this selection is not comprehensive of the more than 52 GFMs released to date [42].

In addition to the three foundational models, we selected the state-of-the-art UNet [33] and Attention UNet [34] models, given their wide applicability in flood mapping tasks [8], [9], [10]. These models have consistently shown strong performance, particularly on the Sen1Floods11 dataset, outperforming all tested GFMs even in limited data scenarios (e.g., using only 10% of the training data) [28], [41]. Therefore, these two models serve as representative examples of state-of-the-art CNN-based methods widely used for flood segmentation. We also included UViT [31], a variant of Prithvi that combines the capabilities of the UNet architecture with the Prithvi encoder using a squeeze-and-excitation layer. The selection of this model was driven by the question, whether channel attention mechanisms can improve flood mapping performance when integrating more than six input channels? Since the original Prithvi encoder is trained on six Harmonized Landsat and Sentinel-2 (HLS) channels, feeding additional channels requires effective channel attention to maintain performance. Finally, we selected TransNorm [32], a simpler Vision Transformer trained from scratch, to evaluate how such models perform compared to large pretrained encoders and state-of-the-art CNNs. This comparison helps assess the trade-offs between model complexity, pretraining, and performance in flood mapping tasks.

TABLE II
SUMMARY OF LARGE GEO-FOUNDATIONAL MODELS (GFMS), INCLUDING PRETRAINING DATA SIZE (IMAGE CHIPS IN MILLIONS), H-R-S INDICATE INCLUSION OF HIGH RESOLUTION IN PRETRAINING, GEO-BENCH [29] AND PANGEA [28] MIOU SHOW SEGMENTATION RESULTS.

| Model | size (M) | H-R-S | Geo-Bench | PANGEA |
|---|---|---|---|---|
| Clay 1.5 [22] | 70 | ✓ | NA | NA |
| DOFA [24] | 8M | ✓ | 0.68 | 0.53 |
| Prithvi 2.0 [19] | 4.2 | × | 0.72 | NA |
| TerraMind [41] | 9 | × | NA | 0.59 |
| Presto [20] | 21 pixels | × | NA | NA |
| Galileo* [25] | 0.13 | × | 0.60 | 0.60 |
| DINO [26] | 1M | × | 0.70 | 0.52 |
| CROMA [44] | 3M | × | NA | 0.55 |
| MOCO [27] | 1M | × | NA | NA |
| Satlas [43] | NA | × | 0.68 | 0.51 |
| ScaleMAE [23] | 0.36 | × | 0.62 | 0.49 |
| Panopticon [45] | ~2.3 | ✓ | 0.53 | NA |

### C. NASA IBM Prithvi Foundational Model

To fine-tune Prithvi 2.0, we used the Terratorch library [46], which allows direct access to pretrained weights. For PlanetScope, the original images were resized to (896×896) using bilinear interpolation as 1024×1024 is not divisible by the patch size (14×14) of the Prithvi 2.0 600M model [19]. This resizing ensures consistent training and validation without padding or cropping. Similarly, Sentinel-1 and Sentinel-2 images were resized to 224×224. Data augmentation methods, such as random vertical and horizontal flips, were applied across all sensors to increase data variability. The original Prithvi 2.0 decoder used during pre-training is based on a Masked Auto Encoder (MAE), which is not ideal for downstream segmentation tasks. Therefore, we used a Feature Pyramid Network (FPN) based UPerNet (Unified Perceptual Parsing Network) decoder for semantic segmentation (Fig. 2). This decoder consists of four deconvolutional layers with activation functions and layer normalization, to map the embedding to the original output. The decoder uses top-down architecture with lateral connections to fuse high- level semantic information into middle and lower levels with minimal additional cost (Fig. 2). Since the Prithvi encoder is trained only on six HLS bands (Blue, Green, Red, NIR, SWIR-1, SWIR-2), appropriate backbone bands need to be selected during fine-tuning. In our experiments, we selected four bands Blue, Green, Red, and NIR for fine-tuning with PlanetScope imagery. For Sentinel-2, we used all six bands (Blue, Green, Red, NIR, SWIR-1, and SWIR-2), and for Sentinel-1, we selected two backbone bands (Blue and Green), aligning with the bands used during Prithvi's pretraining. This band selection ensures compatibility with the pretrained encoder and helps maintain model performance across different sensors.

### D. Clay Foundational Model

The Clay architecture is inspired by Segformer and is based on a Masked Auto Encoder (MAE) ("Made with Clay," 2025). This is the largest GFM model with pretrained over ~70 million chips consisting of different sensors and platforms (e.g., Sentinel-1, Sentinel-2, National Agriculture Imagery Program (NAIP), Moderate Resolution Imaging Spectroradiometer (MODIS), Landsat-8, and Toitū Te Whenua Land Information New Zealand (LINZ). It consists of a Vision Transformer (ViT) encoder to extract hierarchical information from satellite imagery and a convolutional-based decoder to produce segmentation masks. The encoder takes as input a data cube consisting of pixel values with dimensions batch size (B), channels (C), height (H), and width (W), along with auxiliary metadata such as time, latitude and longitude, ground sampling distance (GSD), and waves (central wavelengths).Since our input dataset does not contain time information, we set it to one and GSD of each sensor and central wavelength if each input band is provided as metadata. The model divides the input images into non-overlapping patches of 8×8. The model utilizes multi-head self-attention to capture long-range dependencies and a feed-forward network for feature transformation. The encoder has vector dimensions of 1024, with 24 transformers blocks and 16 attention heads, producing an output feature map of shape (batch, sequence length, and embedding dimension). The decoder segmentation head reshapes the encoder output to spatial feature maps of [batch, embedding dimension, height, and width]. The reshaped features are passed through three 2D convolutional layers with ReLU activation function, with a 3×3 convolution being the last layer providing the final output from the model (Fig. 3). It is noteworthy that in fine-tuning the Clay model, we used the original size of PlanetScope (PS) data (1024×1024). For Sentinel-1 and Sentinel-2 images, the size was set to 320×320 to align with the 8×8 patch size of the model.

### E. Dynamic One-For-All (DOFA) Foundational Model

DOFA foundational model is unique in its category in terms of single versatile Transformers trained on remote sensing data originating from five different sensors (e.g., Sentinel-1, Sentinel-2, Gaofen-2, NAIP, and EnMAP) [24]. While other GFMs (Prithvi, Clay, Scale-MAE, and SatMAE) have a fixed number of input channels, DOFA offers unique advantages by adaptively integrating any number of channels originating from multiple modalities. Thereby, DOFA harnesses the vast potential offered by the fusion of multi-source remote sensing data. The key attribute of DOFA is wavelength-conditioned dynamic patch embedding across remote sensing data to perform uniform representation of multimodal datasets. We utilize DOFA-Large V1.0, which uses 24 transformer blocks, followed by a multilevel neck that builds multiscale feature representation. In the decoder part

UperNet architecture is used which integrates pyramid pooling and feature pyramid fusion to decode these features into semantic segmentation masks. Additionally, an auxiliary FCN head provides intermediate supervision to improve training convergence.

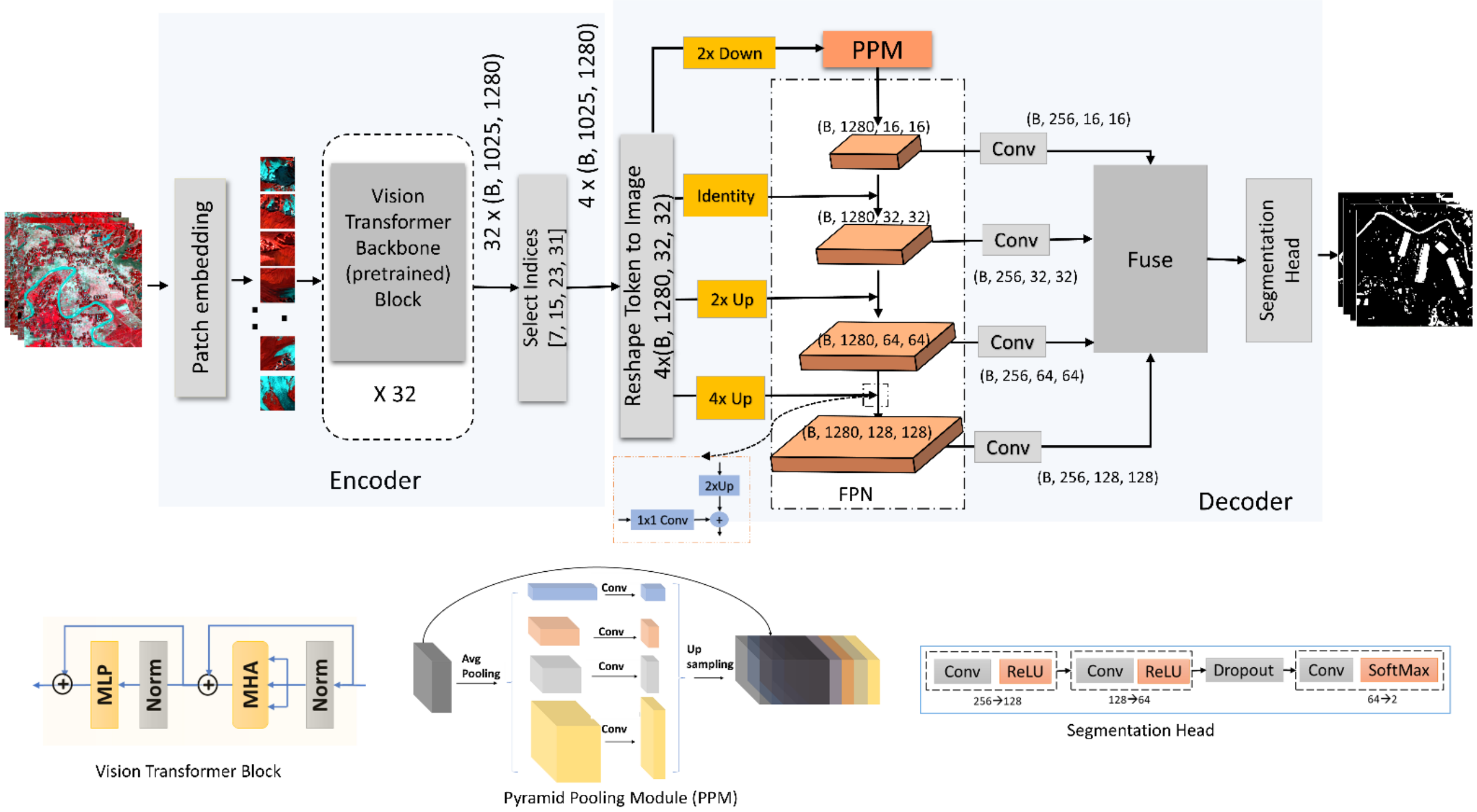

**Fig. 2.** Schematic representation of using Prithvi 2.0 GFM combining UperNet Decoder for flood inundation mapping. The pretrained encoder consists of 32 transformer blocks. Each block applies LayerNorm, multi-head self-attention, and an MLP with Gaussian Error Linear Unit (GELU) activation, producing a sequence of 1280-dimensional feature vectors per patch. These outputs are transformed into 2D shapes using neck modules and passed to a decoder to obtain segmentation mask.

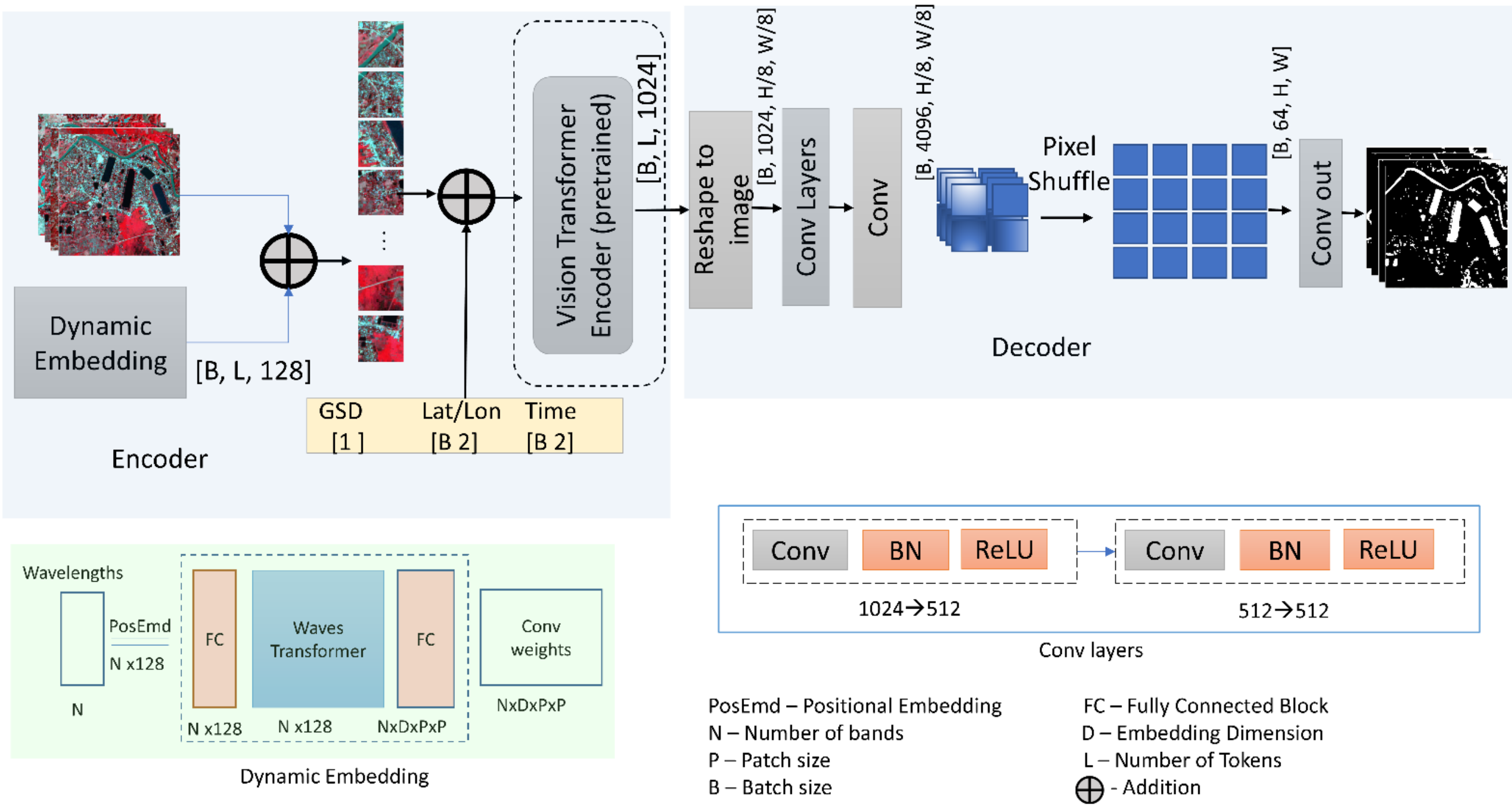

**Fig. 3.** Schematic representation of utilizing Clay GFM for flood inundation mapping. The encoder uses dynamic embedding blocks to create patched from multiband input data and ViT backbone with 16 blocks. In the Decoder project encoder

embedding, reshape patches and convolutional layers to generate segmented flood maps.

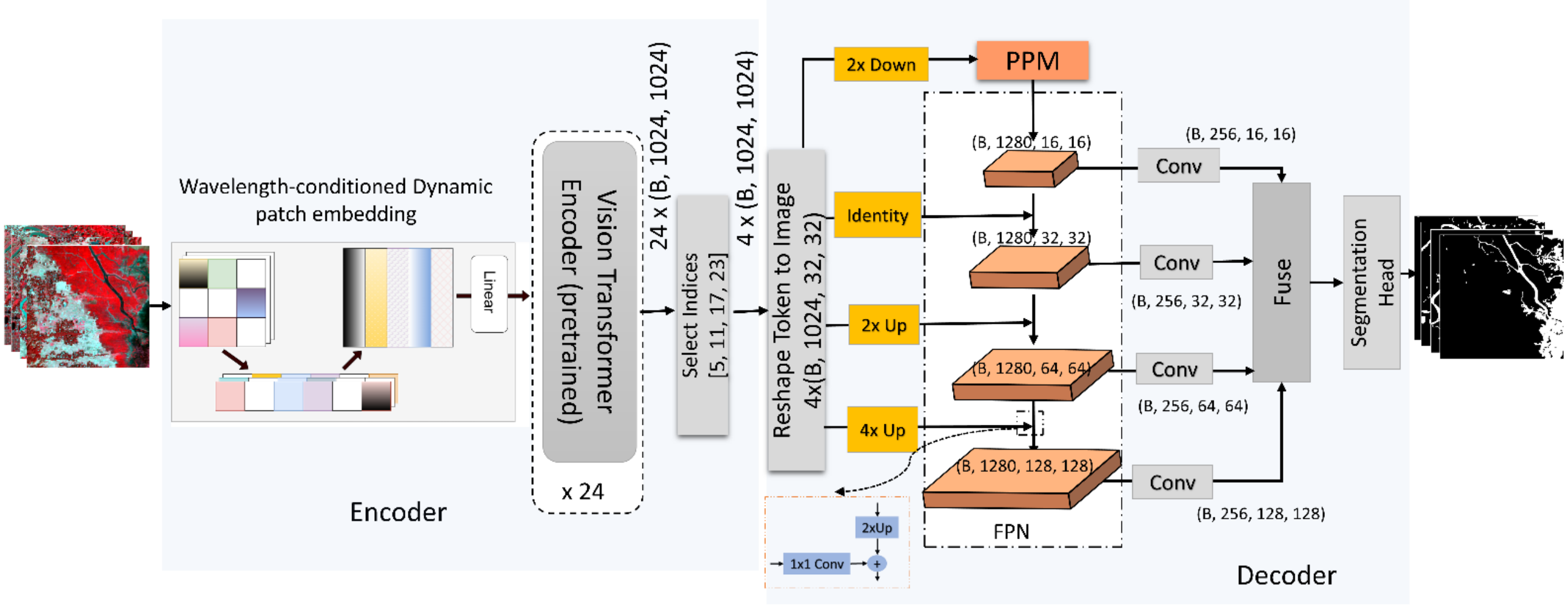

**Fig. 4.** Schematic representation of fine-tuning DOFA GFM for flood inundation mapping. The encoder block uses wavelength conditioned dynamic patch embedding and 24 ViT blocks. In the Decoder part UperNet with FCN head generate segmented flood maps.

### *F. Comparative Models*

We selected several deep learning models for comparative evaluation, including UViT [31], TransNorm [32], U-Net [33], Attention U-Net [34]. U-ViT architecture incorporates Enhanced Squeeze-and-Excitation (SE) blocks to fuse multi-sensor data through dynamic channel-wise attention, along with a U-shaped encoder-decoder structure featuring CNN modules [31]. It leverages a pre-trained Prithvi encoder as a feature extractor and includes three upsampling and CNN layers in the decoder to generate segmentation masks. U-ViT combines the strengths of CNNs for capturing local details and Vision Transformers (ViTs) for modeling global (long-range) dependencies. Its selection is motivated by its flexibility to fine-tune the Prithvi encoder with any number of input channels unlike the original implementation by [19] which only supports up to six channels. Additionally, U-ViT has shown strong performance in glacial lake mapping, a task closely related to flood inundation mapping from satellite imagery. We evaluated U-ViT using both Prithvi 1.0 (100M parameters) and Prithvi 2.0 (600M parameters).

TransNorm [32] is a Transformer-based U-Net architecture designed for biomedical image segmentation. It integrates Transformer modules into both the encoder and the skip connections of the U-Net. The encoder primarily consists of convolutional neural networks (CNNs) to capture semantic and local features, while a parallel Transformer branch encodes long-range dependencies. Additionally, a spatial normalizer is introduced to adaptively recalibrate the skip connection pathways. The decoder, also CNN-based, generates the final segmentation mask. We evaluated the TransNorm model on a flood inundation segmentation task to assess the performance of a Transformer-based U-Net that leverages long-range dependencies, high-level CNN feature extraction, and dual-level attention gates (channel and spatial attention). The model was trained from scratch and compared to large general foundation models (GFMs). For testing, we retained the original input image sizes: 1024×1024 for PlanetScope and 320×320 for Sentinel-1 and Sentinel-2.

U-Net is a widely used deep learning segmentation architecture for flood inundation mapping using Sentinel-1, Sentinel- 2 [8], [9], [47], and Planetscope [48]. Additionally, we include Attention UNet, an enhanced version of UNet that incorporates attention mechanisms to focus on the most relevant features of the input. This modification has been shown to improve performance over the original U-Net in various applications [10]. DeepLabv3+ is one of the leading segmentation models widely used in remote sensing data. Its successful implementation for flood mapping using high resolution unmanned aerial system imagery is also demonstrated [49]. These three models were selected as representative of traditional CNN-based segmentation approaches.

### *G. Experimental Design*

All experiments were conducted using an NVIDIA RTX A6000 GPU. Except for the leave-one-region-out experiments, the dataset was randomly split into 75% training, 10% validation, and 15% testing. All reported results are based solely on the test data. To ensure comparability across experiments, we fixed the random seed to 42, used a batch size of 2 (to accommodate large general foundation models and high-resolution 1024×1024 images), and employed focal loss as the loss function. The AdamW optimizer was used with a learning rate of $1 \times 10^{-4}$. Additionally, a cosine learning rate scheduler with warm-up and restart was applied to facilitate model convergence, with early stopping patience set to 20. We would like to emphasize that in our study, we used all three Geo-Foundational Models (GFMs) with their default decoder architectures example given for semantic segmentation tasks. For example, DOFA [24] and Prithvi [19] utilize the UperNet decoder, while Clay GFM employs a simpler structure consisting of convolutional layers with ReLU activation functions. We did not make any modifications to the models' architecture. Notably, the default implementation of Clay GFM keeps the encoder frozen, unlike DOFA and Prithvi, which allow for fine-tuning of segmentation head keeping clay encoder frozen.

We adhered strictly to these default configurations in our experiments to ensure a fair and consistent comparison. As part of our leave-one-region-out cross-validation experiments, we selected five independent test sites Spain, Ghana, Somalia, Cambodia, and Bangladesh, to represent a diverse range of landscape characteristics and flood challenges. For each experiment, one region (e.g., Spain) was entirely excluded from the training and validation datasets and used solely as an independent test site. This setup allowed us to rigorously evaluate the models' generalization capabilities across heterogeneous geographic and ecological conditions relevant to flood mapping. The same procedure was repeated for each of the five test sites, ensuring a robust assessment of model performance in unseen and varied environments. To conduct few-shot experiments, we divided the training data into smaller subsets and evaluated the models' performance under limited data conditions. In these experiments, we ensured that each subset included at least one image for validation and used a maximum of 10% of the original training data. This setup allowed us to assess the models' generalization capabilities and robustness in data-scarce scenarios.

### *H. Evaluation metrics*

The model's performance is evaluated using two widely accepted metrics for segmentation tasks: mean Intersection over Union (mIoU) and the F1 score, also known as the Dice Similarity Coefficient (DSC). Intersection over Union (IoU) measures the exact overlap between the ground truth and the model's prediction (1), while the DSC represents the harmonic mean of precision and recall (2).

$$mIoU = \frac{1}{k}\sum_{i=1}^{k}\frac{TP}{TP+FP+FN} \quad (1)$$

$$F1 = 2 \times \frac{Precision \times Recall}{Precision + Recall} \quad (2)$$

Where, TP = True Positive, FP = False Positive, FN = False Negative.

## III RESULTS

### *A. Model performance across sensors*

The results of our first experiment using randomly spatially distributed testing sites (Fig. 1) show that GFMs outperform traditional CNNs and the Vision Transformer (TransNorm) trained from scratch (Table III & Fig.5). In the case of PlanetScope, Clay performed slightly better than other models with an mIoU of 0.79, comparatively DOFA and Prithvi 2.0 achieve mIoU of 0.78 and 0.75, respectively. The other variants of Prithvi, UViT600M and UViT100M, also demonstrated competitive performance with mIoU of 0.72 and .67 (Table III), respectively. Our results also highlight the satisfactory performance of the Vision Transformer (TransNorm), which achieved an mIoU of 0.67 compared to traditional CNN architectures (DeepLabv3+: 0.68 UNet: 0.62 and Atten UNet: 0.64). These findings are significant considering previous studies that emphasize the large data requirements for vision transformers and suggest using CNN models for low and moderate data scenarios. Similarly, for Sentinel-2, Clay shows comparatively better performance with an mIoU of 0.72, compared to DOFA's 0.68 and Prithvi 2.0's 0.64. The performance of UViT100M is slightly lower with an IoU of 0.68, followed by TransNorm at 0.66 and UViT600M with 0.60 (Table III). DeepLabv3+ performed slightly better from CNN models, achieving an mIoU of 0.69, compared to 0.67 for U-Net and 0.68 for Attention U-Net. Experiments with Sentinel-1 show superior performance by Prithvi (MIoU=0.57) over Clay and DOFA with mIoUs of 0.51, and 0.45 respectively (Table III). In this experiment, all other models show competitive performance, with UNet achieving an IoU of 0.48 and Attention UNet and DeepLabv3+ slightly higher 0.50. and 0.51 respectively (Table III).

TABLE III
COMPARISON OF GFMS WITH VISION TRANSFORMER AND CNN MODELS ACROSS VARIOUS SENSORS. NUMBERS IN BRACKETS SHOW STANDARD DEVIATION. HIGHER MIOU BETTER PERFORMANCE.

| Model | PlanetScope | | Sentinel-2 | | Sentinel-1 | | Average |
|---|---|---|---|---|---|---|---|
| | F1 | mIoU | F1 | IoU | F1 | mIoU | mIoU↑ |
| Prithvi 2.0 | 0.87 | 0.75(0.14) | 0.78 | 0.64 (0.16) | 0.73 | **0.57** (0.16) | <u>0.65</u> |
| Clay 1.5 | 0.88 | **0.79** (0.12) | 0.83 | **0.72** (0.18) | 0.67 | 0.51 (0.18) | **0.67** |
| DOFA | 0.87 | 0.78 (0.15) | 0.81 | 0.68 (0.16) | 0.62 | 0.45 (0.16) | 0.64 |
| UViT 100m | 0.81 | 0.67 (0.12) | 0.81 | 0.68 (0.16) | 0.66 | 0.49 (0.15) | 0.61 |
| UViT 600m | 0.84 | 0.72 (0.12) | 0.74 | 0.60 (0.18) | 0.67 | 0.50 (0.15) | 0.61 |
| TransNorm | 0.81 | 0.67 (0.12) | 0.78 | 0.66(0.17) | 0.67 | 0.50 (0.17) | 0.61 |
| U-Net | 0.76 | 0.62 (0.20) | 0.80 | 0.67 (0.18) | 0.65 | 0.48 (0.17) | 0.59 |
| Atten-UNet | 0.79 | 0.64 (0.15) | 0.81 | 0.68 (0.20) | 0.67 | 0.50 (0.15) | 0.60 |
| DeepLabv3+ | 0.81 | 0.68 (0.14) | 0.82 | 0.69 (0.16) | 0.68 | 0.51 (0.15) | 0.62 |

Further analysis of the variance within model performance also demonstrates the stable and consistent behavior of different deep learning models. For the PlanetScope sensor, our results show that Clay GFM performs slightly better, with a standard deviation of 0.12, indicating consistent performance across various globally sampled sites (Fig. 1 & Fig. 5). In comparison, the next best-performing model, DOFA, exhibits a slightly higher standard deviation of 0.15, suggesting greater variability in its performance (Table III & Fig. 5). Traditional CNN models such as U-Net and Attention U-Net show even higher variability, with standard deviations of 0.20 and 0.15, respectively, reflecting lower generalizability across test sites. For Sentinel-2 and Sentinel-1, we observe the lowest variance in performance from DOFA and Prithvi, both with a standard deviation of 0.16, compared to Clay GFM, which shows a slightly higher standard deviation of 0.18 (Table III and Fig. 5).

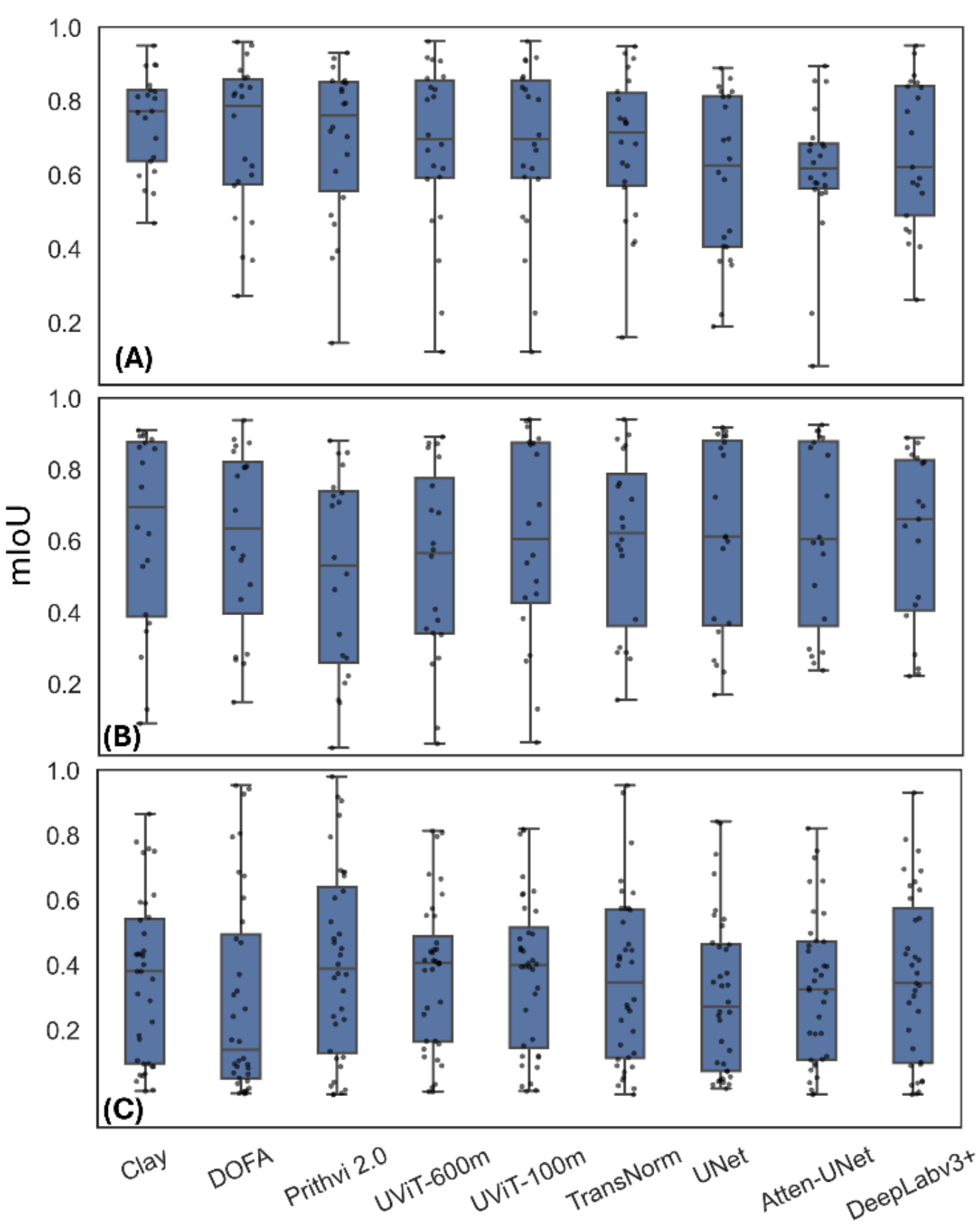

**Fig. 5.** Evaluation of model's performance variation within test dataset (shown in figure 1) for (A) PlanetScope, (B) Sentinel-2, and (C) Sentinel-1.

Overall mean Intersection over Union (mIoU) across sensors and the computed rankings place Clay GFM as the leading model among all evaluated deep learning models, but not by a large amount over other GFMs. Notably, the three major GFMs (i.e., Clay, Prithvi, and DOFA) consistently rank as the top performing models, demonstrating highly competitive performance among the three. These three models are further compared in detail through additional experiments,including computational time analysis, few-shot learning evaluations, and a leave-one-region-out experiment conducted on five selected sites. The qualitative analysis of model predictions reveals that all models are capable of capturing the overall pattern of flooded areas (Fig. 6). However, visual inspection shows that accurately mapping fine details such as non-flooded streets and small land parcel remains a challenge. Fig. 6 presents several examples where Clay GFM outperforms other models in retaining and capturing these fine details, particularly in distinguishing narrow non-flooded streets. It also highlights instances of false positives by other models, where Clay GFM more precisely delineates the actual flooded regions.

### *B. Model's performance in diverse hold-out-test sites.*

To assess the model's spatio-temporal transferability and ensure that our results are not highly influenced by spatial auto-correlation biases, we conducted a leave-one-region-out cross-validation experiment with all three GFMs across all three sensors. To conduct this experiment, we selected five independent test sites Spain, Ghana, Somalia, Cambodia, and Bangladesh, representing diverse eco-regions, physical settings, and mapping challenges. Spain, as an independent test site, highlights the challenges of flood mapping in dense urban areas affected by torrential rainfall. Cambodia presents flood mapping complexities in both dense urban areas and agricultural fields due to monsoonal rains. Meanwhile, Ghana, Bangladesh and Somalia experience flood events primarily driven by heavy rainfall.

Overall, we observe highly competitive performance among all three GFMs, with a maximum difference of 2 to 5 percent between the best and worst performing models across all sensors. Results show comparatively better performance by Clay achieving a mIoU of 0.72 for PlanetScope, 0.66 for Sentinel-2, and 0.51 for Sentinel-1 across all test sites (Table IV). In comparison, Prithvi also delivered competitive results, with a mIoU of 0.70 for PlanetScope, 0.64 for Sentinel-2, and 0.49 for Sentinel-1 (Table IV). The results of DOFA exhibit lower performance over PlanetScope with an mIoU of 0.65. However, DOFA shows competitive performance with Sentinel-2 and Sentinel-1, with mIoU of 0.64 and 0.50, respectively.

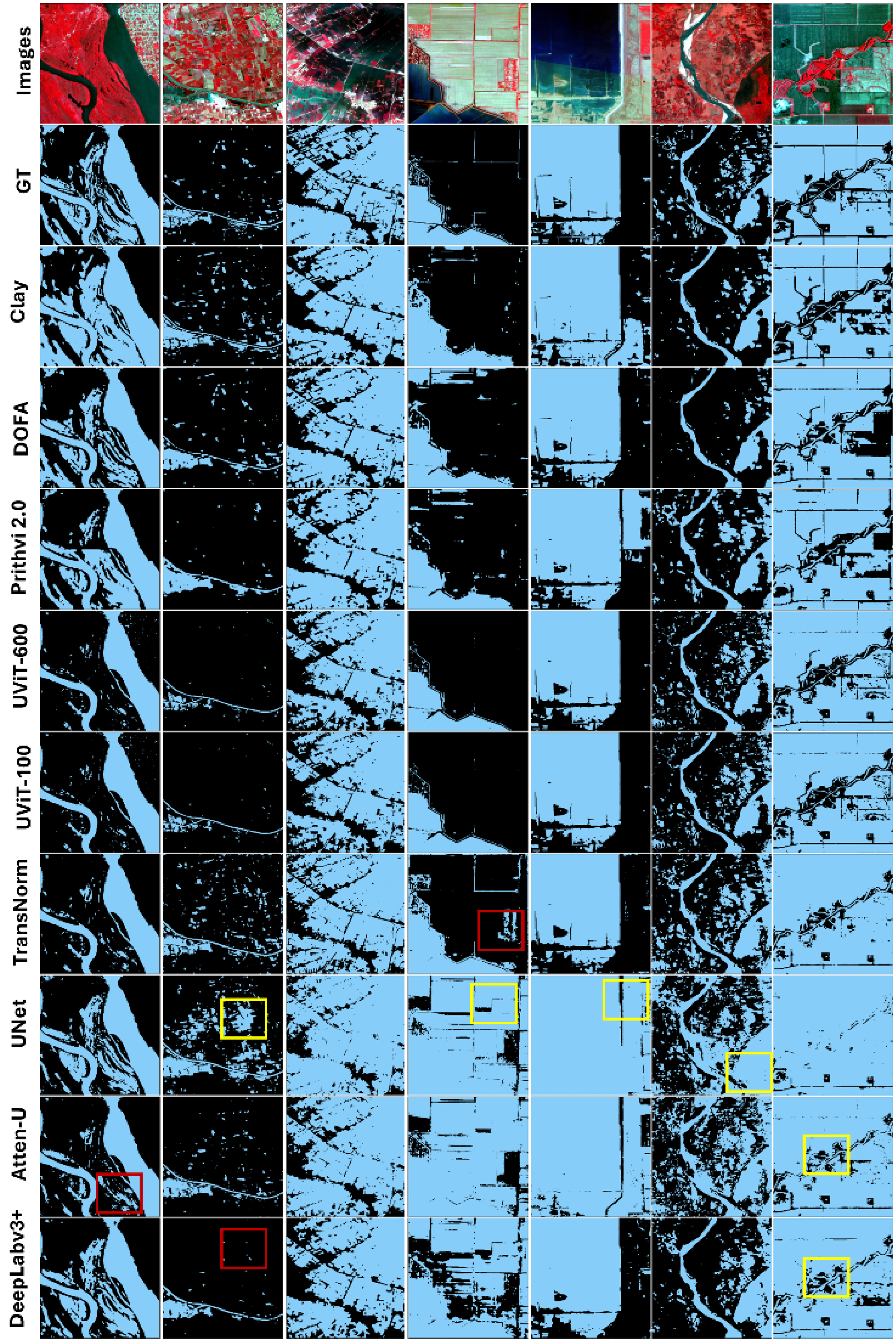

**Fig. 6.** A visual analysis of model performance across spatially distributed test sites covering multiple eco-regions. Results show Clay and Prithvi GFMs outperform traditional CNN models and vision transformers (TransNorm) trained from scratch. Yellow boxes show examples of false positive and red box shows examples of false negative classification by models. (a) Paraguay, (b) and (c) Spain, (d) and (e) Uzbekistan, (f) Bangladesh, and (h) US-Dakota.

Results of the leave-one-region-out experiment for Spain show overall good performance for Clay, achieving an IoU of 0.70 for PlanetScope, 0.58 for Sentinel-2, and 0.53 for Sentinel-1. In contrast, Prithvi 2.0 exhibits somewhat lower mIoUs for PlanetScope and Sentinel-2, with scores of 0.66 and 0.58, respectively, while achieving a marginally higher mIoU of 0.59 for Sentinel-1 (Table IV). We show two examples from Spain: (Fig.7(a)) the best performing site with a highest IoU of 0.80 and (Fig.7(b)) a site where Clay gives a minimum IoU of 0.47. The visual analysis (Fig.7(a)) reveals that Clay effectively captures fine details, even in dense urban areas, compared to Prithvi 2.0 and DOFA, demonstrating its robustness in complex flood mapping scenarios. The visual analysis clearly depicts significant loss of information using Sentinel-1 and Sentinel-2, especially in dense urban regions.

TABLE IV

LEAVE-ONE-REGION-OUT-CROSS-VALIDATION ACROSS FIVE SITES.

| | **PlanetScope** | | **Sentinel-2** | | **Sentinel-1** | |
|---|---|---|---|---|---|---|
| | F1 | mIoU | F1 | mIoU | F1 | mIoU |
| **Spain** | | | | | | |
| Prithvi 2.0 | 0.82 | 0.66 | 0.66 | 0.49 | 0.80 | 0.59 |
| Clay | 0.82 | 0.70 | 0.73 | 0.58 | 0.68 | 0.53 |
| DOFA | 0.79 | 0.67 | 0.76 | 0.62 | 0.64 | 0.50 |
| **Ghana** | | | | | | |
| Prithvi 2.0 | 0.82 | 0.69 | 0.77 | 0.63 | 0.53 | 0.32 |
| Clay | 0.80 | 0.69 | 0.80 | 0.67 | 0.59 | 0.42 |
| DOFA | 0.79 | 0.65 | 0.77 | 0.62 | 0.63 | 0.46 |
| **Somalia** | | | | | | |
| Prithvi 2.0 | 0.90 | 0.78 | 0.87 | 0.75 | 0.57 | 0.40 |
| Clay | 0.88 | 0.79 | 0.86 | 0.76 | 0.61 | 0.44 |
| DOFA | 0.77 | 0.78 | 0.85 | 0.72 | 0.62 | 0.45 |
| **Cambodia** | | | | | | |
| Prithvi 2.0 | 0.85 | 0.71 | 0.82 | 0.66 | 0.82 | 0.66 |
| Clay | 0.80 | 0.73 | 0.80 | 0.69 | 0.77 | 0.64 |
| DOFA | 0.79 | 0.69 | 0.75 | 0.65 | 0.79 | 0.66 |
| **Bangladesh** | | | | | | |
| Prithvi 2.0 | 0.82 | 0.68 | 0.79 | 0.60 | 0.69 | 0.49 |
| Clay | 0.81 | 0.69 | 0.69 | 0.58 | 0.67 | 0.52 |
| DOFA | 0.71 | 0.56 | 0.77 | 0.63 | 0.54 | 0.45 |
| **Sensor wise model's performance** (Average Value) | | | | | | |
| | mIoU (std) ↑ | Rank↓ | mIoU (std)↑ | Rank↓ | mIoU (std)↑ | Rank↓ |
| Prithvi 2.0 | 0.70(0.05) | 2.0 | 0.63(0.09) | 2.2 | 0.49(0.13) | 2.0 |
| Clay | 0.72(0.04) | 1.0 | 0.66(0.07) | 1.6 | 0.51(0.08) | 1.8 |
| DOFA | 0.67(0.07) | 2.6 | 0.65(0.04) | 2.2 | 0.50(0.09) | 2.0 |

Results of Ghana show Clay and Prithvi 2.0 perform competitively well, each achieving IoU scores of 0.69, whereas DOFA achieves an IoU of 0.65 for PlanetScope. For Sentinel-2 Clay shows a slight advantage, with IoU scores of 0.67 compared to Prithvi's 0.63 and DOFA's 0.65, respectively (Table 4). Visual inspection confirms that all three models can handle sparsely covered cloud coverage well (Fig.7(c)). This example also confirms that all three models perform well with Sentinel-1 and Sentinel-2 data, provided the majority of pixels belong to the relevant pixel class (i.e., flood). However, Prithvi faces challenges in accurately capturing small land parcels (Fig.7(c)). The misclassifications (Fig.7(d)) could be primarily introduced by cloud cover in both PlansetScope and Sentinel-2. Additionally, the analysis reveals that while Prithvi 2.0 effectively detects flood regions, it has limitations in mapping smaller scale details compared to Clay (Fig.7(c)).

Cambodia shows both Prithvi 2.0 and Clay perform competitively, with Clay demonstrating a slight advantage by achieving an mIoU of 0.73 compared to Prithvi's 0.71 for PlanetScope. In contrast, DOFA achieves a lower mIoU of 0.69. For Sentinel-2, Clay performs consistently better, achieving an mIoU of 0.69, compared to Prithvi's 0.66 and DOFA's 0.65. In the case of Sentinel-1, Prithvi and DOFA outperform with an mIoU of 0.66 compared to Clay's 0.64. The Bangladesh test site represents flooding conditions due to heavy rain, varying from dense urban regions to open agricultural fields. Evaluation of this site exhibits competitive performance by Clay and Prithvi for PlanetScope, with mIoU scores of 0.69 and 0.68, respectively. In contrast, significantly lower performance is observed for DOFA, with an mIoU of 0.56 (Table IV). Unlike other sites, DOFA outperforms the other models for Sentinel-1, with an mIoU of 0.63 compared to 0.53 and 0.60 achieved by Clay and Prithvi.

Visual analysis supports these observations, showing that Clay seems to capture fine details, while Prithvi 2.0 struggles with smaller-scale flood features (Fig. 7(i)). Overall, the average ranking indicates that Clay performs better as a GFM compared to other models when using PlanetScope and Sentinel-2 data. However, we observe a tie in the average ranking for Sentinel-1. The results of leave- one-region-out cross-validation show highly competitive numbers, with marginally higher performance demonstrated by Clay over other GFMs. These results might introduce ambiguity in model selection for end users. To address this, we included all results for sensors and test sites, showing that Clay has a higher mIoU in four out of five sites (Fig. 8 (a)). This analysis indicates the slight advantage of using the Clay GFM over other models. We also conducted sensor-wise analysis, which indicates that Clay outperforms in four out of five sites for PlanetScope (Fig. 8(b)), shows better performance in three out of five sites for Sentinel-1 (Fig. 8(c)), and out- performs in four out of five sites for Sentinel-2 (Fig. 8(d)).

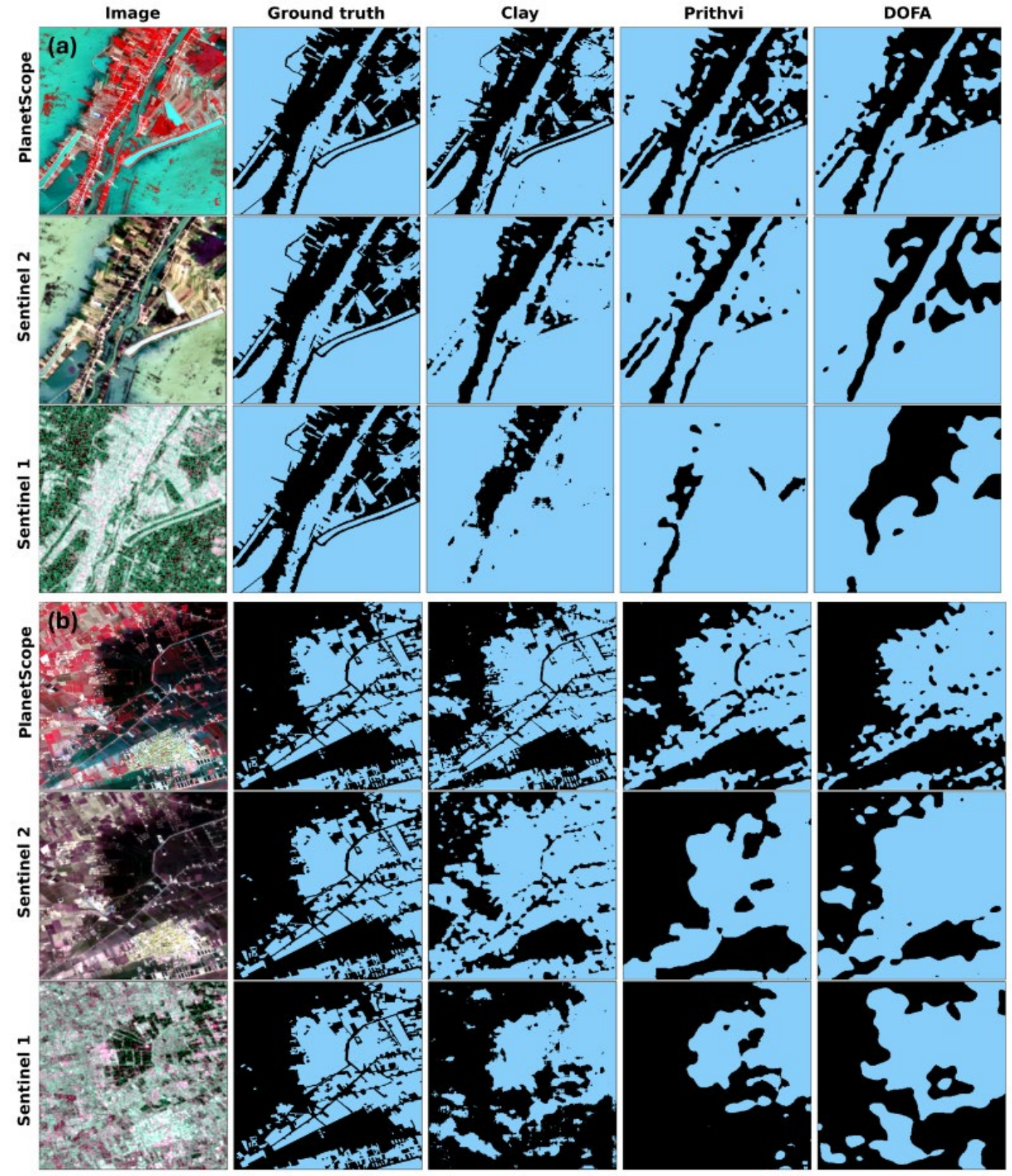

**Fig. 7.** Comparison of Clay, DOFA, and Prithvi GFMs across PlanetScope, Sentinel-2, and Sentinel-1 in the leave-one-region-out experiment, examples are shown from (a) Cambodia and (b) Spain.

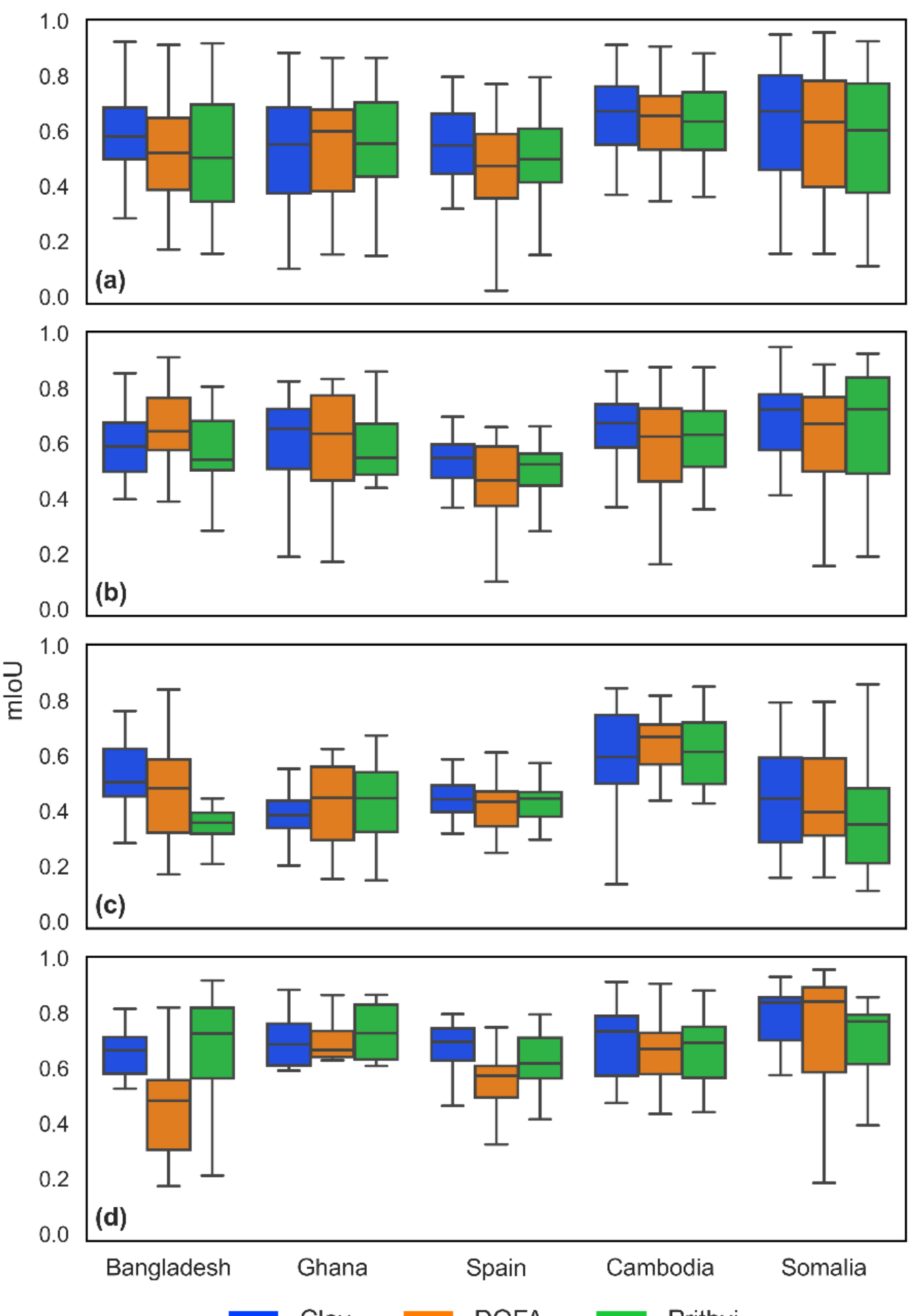

**Fig. 8.** The model's performance over different hold-out regions, (a) aggregated across all sensors, shows relatively better performance of Clay in four out of five regions. (b) PlanetScope, (c) Sentinel-1 and (d) Sentinel-2.

### *C. Few-shot experiments and Computational efficiency*

The evaluation over PlanetScope data reveals impressive results by Clay, achieving an mIoU of 0.64 with just 5 images as training data, compared to 0.35 and 0.24 by DOFA and Prithvi 2.0, respectively (Fig.9(a)). In case of PlansetScope we observe after approximately 50 input images, the performance of Clay largely plateaus. In contrast, Prithvi and DOFA continue to improve as additional input images are incorporated, with slight gains observed even up to 150 input images. These trends indicate that larger models require more training data to fully exploit their representational capacity, particularly Prithvi (650M trainable parameters) and DOFA (410M trainable parameters).

It is important to note that our dataset includes large tile sizes, meaning a subset of 5 tiles of size 1024×1024 translates to 80 smaller tiles of 256×256. To further investigate, we fine-tuned the models using only two training tiles. Clay achieved an mIoU score of 0.58, while Prithvi and DOFA struggled with scores of 0.15 and 0.17, respectively (Fig.9(a)). The results revealed that Clay substantially outperforms Prithvi in data-limited scenarios. Similar results are observed for the Sentinel-2 dataset, where Clay achieved an impressive mIoU of 0.59 with just 5 training images, compared to Prithvi's 0.47 and DOFA's 0.46.

In the case of Sentinel-2, we observe relatively better performance of DOFA when trained on only two images, showing an mIoU of 0.45, whereas Clay and Prithvi achieved only 0.37 and 0.23, respectively (Fig. 9(b)). Visual analysis of the few-shot experiments supports the superior performance of Clay GFM compared to Prithvi and DOFA. As shown in Figure 10, Clay consistently produces reliable flood maps with fine-tuning on just 2 or 5 images, whereas Prithvi and DOFA struggle significantly under the same conditions. These findings demonstrate that in scenarios with very limited training data, Clay is the superior choice, making it particularly well-suited for applications where labeled data is a constraint. Clays' ability to fine tune quickly is a predictable result- given its much smaller model size- and demonstrates that in label limited situations, lighter models with less parameters to fine-tune may be a good choice. We attribute Clay's success in data-limited scenarios to its strategic use of diverse training data. The model benefits from a combination of high-resolution drone imagery (NAIP), which provide detailed spatial information crucial for flood mapping. In contrast, Prithvi 2.0 relies on a massive yet moderate-resolution (30m) HLS dataset, which integrates six spectral bands common to both Sentinel and Landsat sensors. While Prithvi's approach ensures broad spectral coverage, it lacks the fine spatial details that Clay leverages from higher-resolution sources. This difference in training data composition likely contributes to Clay's superior performance, makes it efficient, and lighter, and even with smaller training sets can map fine details.

The leave-one-region-out cross-validation experiments compared the three GFMs, with Clay slightly outperforming Prithvi 2.0 and DOFA in terms of mIoU and average ranking. However, in several cases, all models demonstrated highly competitive performance, which may introduce some ambiguity for end users when selecting the best model. To address this, we compared the training time of all three GFMs using identical datasets (311 training and 35 validation images) and using the same hardware setup (NVIDIA RTX A6000 GPU) to ensure a fair and comparable evaluation of training efficiency. The results reveal Clay's computational efficiency, taking approximately 1.6 minutes per epoch, whereas Prithvi 2.0 requires approximately six minutes per epoch and DOFA takes approximately three minutes making Clay nearly three times faster than Prithvi and twice as fast as DOFA (Fig.9(c)). These results were expected theoretically, as Prithvi 2.0 has 650M trainable parameters, DOFA has 400 trainable parameters, in comparison Clay has only 26M trainable parameters. Our results indicate that heavy and complex models do not always offer optimum solutions. These findings are particularly important for large-scale datasets and scenarios with limited computational resources, as a faster model allows for more iterations and optimization within the same time frame, and cheaper costs.

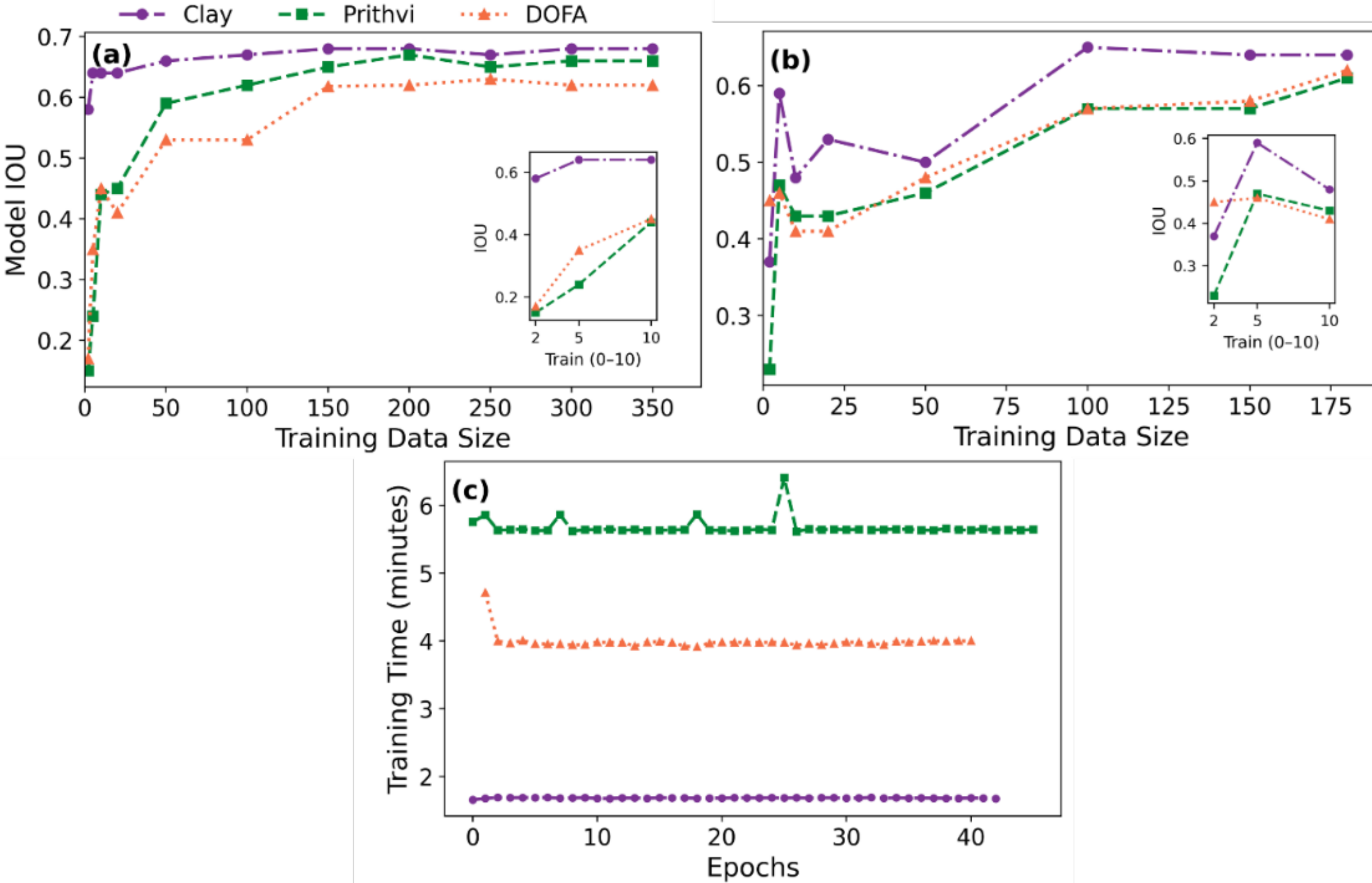

**Fig. 9.** Assessment of Model Performance in few-shot experiments and computational Time. (a) Few-Shot experiment using PlanetScope data: Clay model is nearly three times faster than the Prithvi GFM model and nearly twice as fast as DOFA. Few-shot experiments on PlanetScope (b) Few-Shot experiments using Sentinel-2 show superior performance by Clay with minimal data compared to Prithvi and DOFA. (c) shows computational time for all three models for fine-tuning using the same (Planetscope).

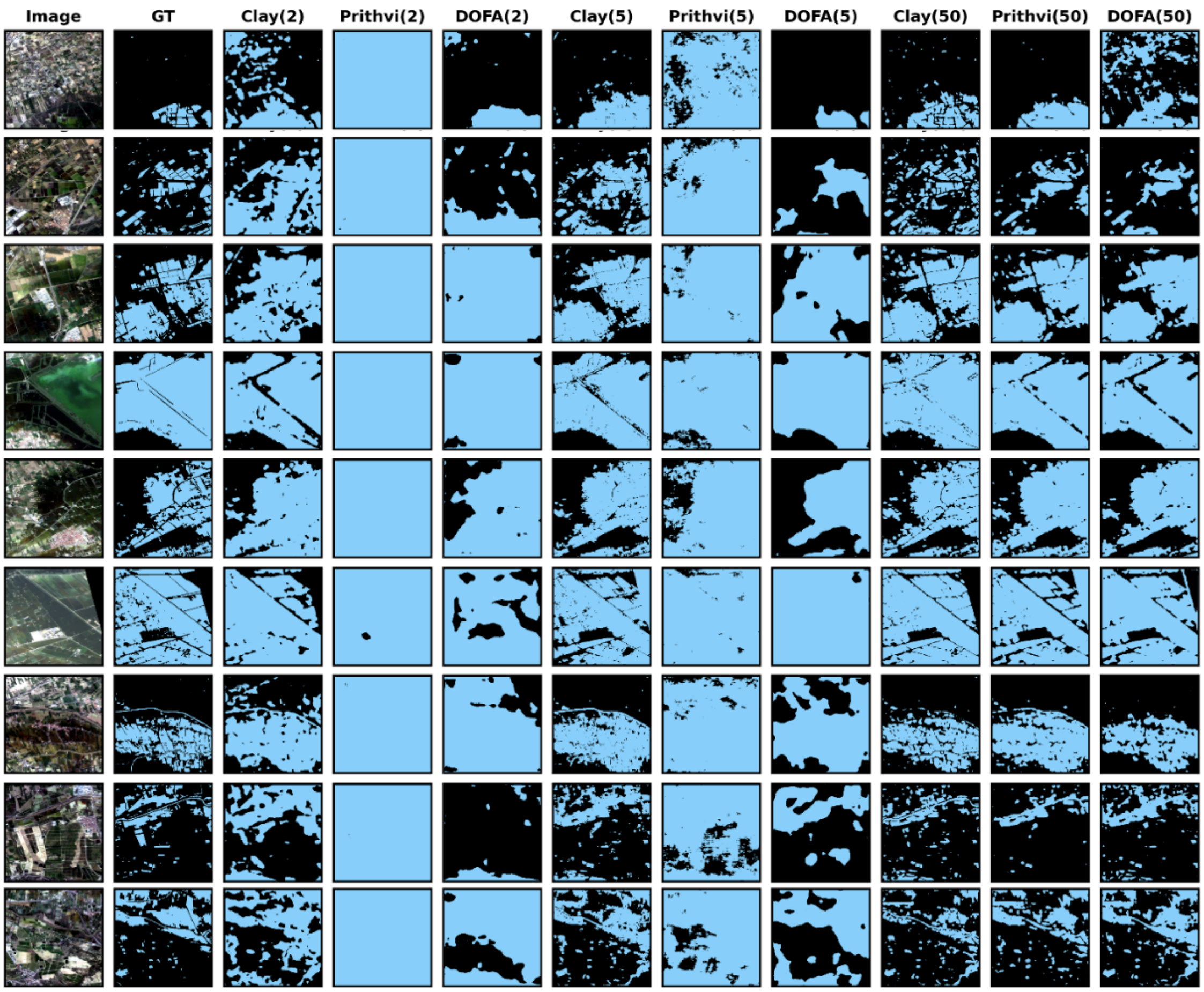

**Fig. 10.** Few-shot experiment results for Clay, Prithvi, and DOFA using PlanetScope data. The numbers in parentheses indicate the number of images used for fine-tuning each model. Visual analysis revealed that Clay GFM outperformed the others, even with the least amount of training data (e.g., 2 and 5 images).

## IV Discussion

### *A. What Value Do Geo-Foundational Models Bring Compared to Traditional CNNs?*

This study is strongly motivated by the recent advancements in Geo-Foundational Models (GFMs) and their demonstrated ability to encode temporal and location embeddings, which are critical for improved flood inundation mapping. The use of large pretrained encoders built on advanced Vision Transformer (ViT) architecture was expected to significantly enhance the flood mapping domain. However, the results on the Sen1Floods11 benchmark dataset [30] revealed that all tested GFMs underperformed compared to baseline U-Net, even in limited data scenarios [28]. These observations raise an important question within the scientific community: Do GFMs truly advance flood mapping compared to traditional CNN approaches? However, we would like to highlight that the evaluation protocol used in the PANGEA benchmark with Sen1Floods11 keeps the model's pretrained weights frozen, thereby leaving the full potential of GFM with fine-tuning yet unlocked.

To address this, we conducted a comprehensive evaluation by testing three of the largest GFMs against traditional CNN-based models (UNet, Attention UNet and DeepLabv3+) and Vision Transformers trained from scratch using high quality label dataset FloodPlanet [35]. These models were assessed across three widely used satellite sensors: Sentinel-1, Sentinel-2, and PlanetScope. Our results demonstrate the advantages of using massive pretrained encoders for flood mapping across different satellite sensors (Table 3). The selected GFMs consistently outperformed traditional CNNs, and Vision Transformers (ViTs) trained from scratch (Table 3 and Figure 6). We observe an approximate 11% increase in mIoU when comparing the best-performing GFM Clay (mIoU 0.79 (0.14) to the best-performing CNN-based DeepLabv3+ (mIoU 0.68 (0.14)). The other GFMs, DOFA and Prithvi, also significantly outperform the CNN models, with DOFA showing 0% and Prithvi an 7% improvement (Table 3).

Interestingly, we observe that UViT, which combines UNet and Prithvi encoder with a squeeze-and-excitation layer, significantly underperforms compared to the Prithvi encoder combines with a UPerNet decoder. In contrast, TransNorm a ViT trained from scratch, outperforms UNet with a 5% increase in mIoU. This highlights the potential of ViT architectures for flood segmentation tasks, even when trained from scratch. Similar observations can be drawn from Sentinel-1, where the best-performing GFM, Prithvi, shows an 7% improvement over Attention U-Net and a 9% increase over U-Net (Table 3). Whereas DeepLabv3+ slightly outperform Clay with 1% and DOFA with 6%.

In contrast, in case of Sentinel-2 we observe that the performance gap in mIoU between GFMs and other models narrows significantly. For example, the best-performing GFM, Clay, shows only a 3% improvement compared to DeepLabv3+ and a 5% improvement compared to U-Net. These observations highlight the importance of GFMs in high-resolution datasets such as PlanetScope and Sentinel-1. However, we infer that Sentinel-2, with its 10 spectral channels at 10 m spatial resolution, provides sufficient spatial context to make traditional models more competitive with GFMs. It is important to note that the labels for all sensors were manually annotated using high-resolution PlanetScope imagery. As a result, moderate-resolution Sentinel imagery may not capture fine-scale details with the same spatial accuracy, which can limit its ability to precisely detect small or narrow flooded features. Thus, we observe the highest difference between GFMs and CNN over PlanetScope data.

The recent findings (e.g., [41]) suggest that TerraMind and other GFMs underperform compared to U-Net in flood mapping. In addition, [50] reported ViT encoders trained from scratch without any pre-training can surpass (IoU 83.11) massively pretrained Prithvi encoders using Sen1Floods11 dataset [30]. However, it's important to note that in those experiments, the GFM encoders were kept frozen to demonstrate their ability to produce fast and efficient results, unlike U-Net, which was trained from scratch. Even with frozen encoders, U-Net only marginally outperformed GFMs by 0.64%. In contrast to these studies, we fully fine-tuned the Prithvi and DOFA backbones to demonstrate their potential to outperform CNNs. Notably, even the frozen encoder of the largest GFM, Clay, outperformed all CNN-based models (Table 3). These results highlight the significant potential of GFMs for flood mapping, either through full fine-tuning when maximum precision is required, or by keeping the encoder frozen when fast inference and stability are the priority.

To further examine this trade-off, we conducted an ablation study of the Clay GFM, comparing frozen-encoder and fully fine-tuned configurations across a range of learning rates (Table 5). The results show that Clay is highly sensitive to learning-rate selection when fully fine-tuned, with model performance (mIoU) varying by up to 51% across learning rates from 1e-3 to 1e-7. Based on these findings, we recommend keeping the encoder frozen for robust and efficient deployment and applying full finetuning only when the highest possible accuracy is required and careful hyperparameter tuning is feasible.

TABLE V

ABLATION STUDY OF CLAY GFM WITH FROZEN AND FULLY FINE-TUNED ENCODER.

| **Encoder** | **Learning rate** | **mIoU** | **F1** |
|---|---|---|---|
| Frozen | 1e-5 | 0.79 | 0.88 |
| Fully fine-tune | 1e-3 | 0.30 | 0.37 |
| Fully fine-tune | 1e-5 | 0.77 | 0.87 |
| Fully fine-tune | 1e-6 | 0.81 | 0.89 |
| Fully fine-tune | 1e-7 | 0.79 | 0.88 |

When directly comparing baseline U-Net results on FloodPlanet [35] reveals Clay GFM shows 4% mIoU improvement. These results further strengthen the potential offered by GFMs in improving flood inundation mapping capabilities compared to traditional CNN. Most notably, Clay GFM exhibited significantly lower variance, with a standard deviation of 0.10 compared to U-Net's 0.22 (Table 6), indicating more generalized and stable performance across diverse eco-regions and site-specific challenges. Similar, advantage of pretrained encoder for flood mapping using Sen1Floods11 dataset is reported by [51], leveraging Prithvi generalization capabilities integrated into U-Net architecture to retain fine details. The proposed model outperforms the baseline U-Net, achieving an IoU score of 87.70, compared to 82.54 by U-Net.

We also conducted a detailed comparative evaluation of three GFMs namely Clay, Prithvi, and DOFA using leave-one-region-out-cross-validation, few-shot learning experiments, and computational efficiency analysis. In cross-validation experiments, Clay showed slightly better (5-2%) performance compared to Prithvi and DOFA (Table 6 and Figure 7 and 8). The difference became more pronounced in few-shot settings: with only 5 training images, Clay achieved an mIoU of 0.64, significantly outperforming DOFA (0.35) and Prithvi 2.0 (0.24) (Fig. 9a). In terms of computational efficiency, Clay, with only 26 million parameters, was found to be approximately 3× faster than Prithvi (650M parameters) and 2× faster than DOFA (410M parameters), making it a highly efficient choice for flood mapping algorithm choice when label dataset in scare. (Fig. 9c).

In addition, we emphasize that the assessment of GFMs should extend beyond model accuracy alone to include spatial transferability, generalization across diverse datasets and domains, and deployment efficiency [52]. These characteristics ultimately make GFMs a more energy-efficient alternative to training task, region, and dataset-specific models from scratch [52]. These sustainable development benchmark [52] further strengthen the value of GFM in real world application compared to traditional approaches.

TABLE VI

COMPARATIVE EVALUATION BASELINE U-NET MODEL WITH CLAY GFM ACROSS ALL 19 TEST SITES USING MIOU IN LEAVE-ONE-REGION-OUT EXPERIMENT. *U-NET RESULTS ARE OBTAINED FROM [35].

| Event name | U-Net* | Clay GFM |
|---|---|---|
| US-Kansas | 0.35 | 0.42 |
| US-Texas | 0.85 | 0.82 |
| Ghana | 0.59 | 0.69 |
| Nepal | 0.82 | 0.84 |
| Cambodia | 0.56 | 0.73 |
| US-Alabama | 0.83 | 0.84 |
| Spain | 0.82 | 0.70 |
| Uzbekistan | 0.37 | 0.63 |
| Nigeria | 0.54 | 0.81 |
| Somalia | 0.80 | 0.79 |
| US-Carolina | 0.81 | 0.70 |
| Bolivia | 0.75 | 0.72 |
| Colombia | 0.79 | 0.73 |
| US-Oklahoma | 0.82 | 0.81 |
| Bangladesh | 0.66 | 0.69 |
| US-Nebraska | 0.57 | 0.63 |
| US-Dakota | 0.55 | 0.70 |
| US-Arkansas | 0.91 | 0.86 |
| Paraguay | 0.73 | 0.73 |
| **Mean** | 0.69 | **0.73** |
| **Standard deviation** | 0.22 | **0.10** |

### *B. Limitations and future outlook*

Overall, GFMs demonstrate improved flood mapping capabilities compared to traditional CNNs (e.g., U-Net) on both randomly distributed test sites and leave-one-region-out cross-validation (Table 3 & Table 4). Despite these improvements, several limitations remain, which we categorize as follows: 1) Spctral ambiguity: Challenges persist in distinguishing mixed spectral signatures resulted from shallow water accumulation. (Fig. 11). These challenging mapping conditions often arise in areas with water presence in small farmlands or narrow urban streets, where weak reflectance signals result in ambiguous spectral characteristics (Fig. 11). Figure 11 clearly illustrates these limitations, showing that even the best-performing model, Clay, struggles to accurately map small land parcels with strong spectral similarity. The limitation of model's performance also stems from input dataset, which relies solely on optical imagery, which limits its ability to capture certain flood characteristics (e.g., local topography and flow direction). To address these challenges, we recommend incorporating complementary datasets such as Digital Elevation Models (DEM), slope, and flow direction either sourced externally or generated synthetically to enhance model performance. For example, recent work by [15] demonstrates the effectiveness of integrating such auxiliary data (DEM,slope, Permanent Water, and Height above Nearest Drainage) with satellite data in improving flood mapping accuracy using model ensemble approach. 2) The other two major sources of error in flood mapping remain cloud cover and cloud shadows, which pose significant challenges, especially during active flood events. Misclassification in the presence of cloud cover is evident from Figure 11. In addition, the leave-one-region-out cross-validation results indicate the lowest performance for U.S.–Kansas and Uzbekistan, highlighting limited spatial transferability of the model to these two regions.

To minimize the impact of cloud-related data gaps, future research should focus on leveraging the generative capabilities of recent Geo-Foundational Models (GFMs), such as TerraMind. TerraMind is a modality-agnostic generative model that supports any-modality-to-any-modality data generation. Notably, it has demonstrated promising performance in generating Sentinel-2 imagery from Sentinel-1 data with a mean absolute error (MAE) of 0.07, and vice versa with an MAE of 2.9. These capabilities offer a strong foundation for addressing cloud-induced data gaps. Furthermore, flood mapping results could be enhanced by integrating spatiotemporal gap-filling algorithms, which help maintain consistency and completeness in flood extent predictions. 2) Data limitations: Although the FloodPlanet dataset is one of the most comprehensive, hand-labeled public datasets using high-resolution PlanetScope imagery, it may still contain human annotation uncertainties. These errors primarily arise from ambiguous spectral signatures, particularly in areas of dense urban regions, and regions where flooded surfaces are spectrally similar to wet soil, or shaded regions.

Future direction of research should be inclined toward combining late fusion methods (e.g., [53] to leverage full potential of multi-modal data, advanced CNN (U-Net and DeepLabv3+) methods combined with different GFMs or other pretrained encoder offer a promising direction for flood mapping. Such hybrid models will leverage advantage of capturing local feature extraction strengths of CNNs and massive pre-trained transformers encoders will learn long-range dependencies of multi-source remote sensing data (i.e., optical, SAR, DEM, and Precipitation) to leverage the full potential offered by such diverse data [51]. In this category, DOFA and Clay offer unique opportunities and pragmatic approach to experiment with different combinations of input datasets compared to Prithvi where fixed number of input channels are expected. In addition, the fast-growing availability of different GFMs [25], [41], [54], [55] provides ample scope for evaluation, experimentation, and architectural improvements. We believe that GFMs hold immense potential for application-specific tasks, but most existing work has focused on broad, general-purpose applications.

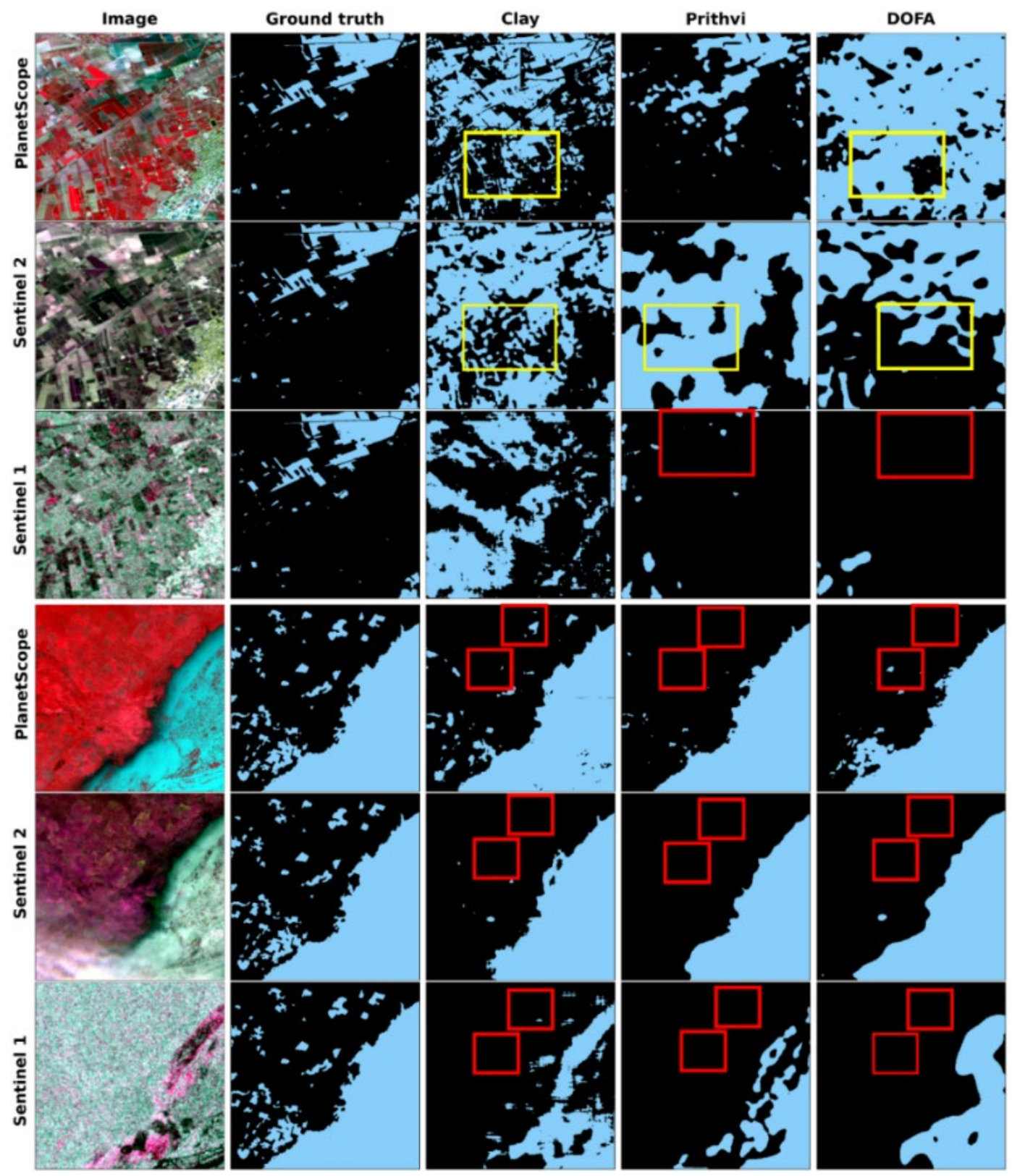

**Fig. 11.** The snapshot of model's limited performance primarily due to similar spectral signature, small land parcels, and cloud cover. Yellow boxes show false positive by different GFMs, and red boxes show false negative classified pixels.

To unlock their full capabilities, future efforts should prioritize the development of task-specific models through extensive experimentation with decoder architecture, efficient fine-tuning strategies, and cross-modality fusion techniques that preserve both local and global information. For example, recent study by [56] highlights improved performance of GFMs with smaller learning rate for efficient fine-tuning of pretrained encoder for semantic segmentation task using moderate and high data regimes (50%-100%).

Additionally, parameter-efficient fine-tuning methods, such as adapter-based approaches that have proven effective in Large Language Models [57] and GFMs [45] should be explored to enhance adaptability and reduce computational overhead. We would like to highlight that the implementation of Prithvi GFM is straightforward and easier to implement from an end-user perspective. Its pretrained weights, encoder, and decoder can be directly accessed through the TerraTorch library, making it easy to use and integrate. We recommend using Prithvi GFM for large-scale flood mapping with widely used public sensors such as Sentinel-1 and Sentinel-2. For site-specific, detailed mapping, especially where fine-grained features (e.g., streets) are a priority, Clay GFM is an excellent choice. It enables the creation of high-resolution flood databases that can support local authorities, urban planners, and policymakers in decision-making and disaster response.

## V CONCLUSION

Our study highlights the efficiency and stability of GFMs compared to traditional CNNs and vision transformers trained from scratch. We selected Clay, DOFA, and Prithvi 2.0—the best-performing models from our test sites, for detailed comparison using leave-one-region-out cross-validation, computational efficiency, and few-shot experiments. The results show competitive performance among GFMs with Clay demonstrating marginal improvement over other models. Visual analysis confirms Clay's effectiveness in capturing fine details and handling diverse environmental conditions across floods and sensor inputs. These findings emphasize the importance of selecting models that balance performance, efficiency, and scalability for flood mapping tasks. Evaluation of computational efficiency, and adaptability with limited training samples reveal slight out performance of Clay compared to other GFMs, and emerge as the more efficient and lightweight model, demonstrating superior adaptability when trained on a small dataset. Its ability to capture fine details stems from extensive pre- training on diverse remote sensing data including high-resolution satellite and drone imagery, making it well applied to flood mapping- although Privthi's competitive accuracy and easier implementation make it also good model choice for end users making flood maps in near real time or over large areas. With the rise of new sensors for inundation mapping, GFMs can contribute to mapping these sensors with minimal labels for fine-tuning. We recommend, as a future direction of research, the integration of diverse input datasets, experimentation with hybrid GFM-CNN architectures, the use of multiple decoders, and efficient fine-tuning techniques such as adapter modules. These strategies have the potential to further advance flood inundation mapping using satellite data.

### Author Contribution

**SK:** Conceptualization; Data curation; Methodology; Formal analysis; Visualization, Software, Writing– original draft, and Writing - review & editing. **LM:** Conceptualization, Methodology, Writing - review & editing. **BT**: Conceptualization; Resources; Supervision; Funding Acquisition; and Writing - review & editing. **GZ:** Conceptualization; Writing - review & editing.

### Code and Data Availability

GitHub https://github.com/Sk-2103/Advancing-Flood-Inundation-Mapping-Using-Geo-Foundational-Models.git

FloodPlanetdata
https://datacommons.cyverse.org/browse/iplant/home/shared/commons_repo/curated/Zhijie_FloodPlanet_2023